\documentclass[runningheads]{llncs}
\usepackage{graphicx}
\usepackage{amsmath,amssymb} % define this before the line numbering.
\usepackage[width=122mm,left=12mm,paperwidth=146mm,height=193mm,top=12mm,paperheight=217mm]{geometry}
\usepackage{color}

\usepackage{algorithm}
\usepackage[noend]{algpseudocode}

\usepackage{subcaption}% http://ctan.org/pkg/subcaption
\DeclareCaptionSubType*{algorithm}

\DeclareCaptionLabelFormat{alglabel}{Alg.~#2}

\newcommand{\fakesubsection}[1]{\smallskip\noindent\textbf{#1:}}

\usepackage{url}
\usepackage{graphicx}
\graphicspath{{imgs/}}
\DeclareGraphicsExtensions{.pdf,.jpg,.png}
\usepackage{xcolor}

\newcommand{\etal}{\emph{et al}.}

\begin{document}
	
% \linenumbers
\pagestyle{headings}
\mainmatter
\def\ECCV16SubNumber{473}  % Insert your submission number here

\title{Segmental Spatiotemporal CNNs for\\ Fine-grained Action Segmentation}

%\titlerunning{ECCV-16 submission ID \ECCV16SubNumber}
%\authorrunning{ECCV-16 submission ID \ECCV16SubNumber}

%\author{Anonymous ECCV submission}
%\institute{Paper ID \ECCV16SubNumber}

\authorrunning{C Lea, A Reiter, R Vidal, G Hager}

\author{Colin Lea \hspace{11px} Austin Reiter \hspace{11px} Ren\'{e} Vidal \hspace{11px} Gregory D. Hager}
\institute{	Johns Hopkins University\\ 
%	Baltimore, MD, USA \\
%	 clea1@jhu.edu, areiter@cs.jhu.edu, rvidal@cis.jhu.edu, hager@cs.jhu.edu}
	 \{clea1@, areiter@cs., rvidal@cis., hager@cs.\}jhu.edu}

%\institute{Department,\\
%	University\\
%	\email{ \{author1,author2\}@univ.edu}
%}

\maketitle

% As a general rule, do not put math, special symbols or citations in the abstract or keywords.
\begin{abstract}

Joint segmentation and classification of fine-grained actions is important for applications of human-robot interaction, video surveillance, and human skill evaluation. 
However, despite substantial recent progress in large-scale action classification, the performance of state-of-the-art fine-grained action recognition approaches remains low. We propose a model for action segmentation which combines low-level spatiotemporal features with a high-level segmental classifier. 
Our spatiotemporal CNN is comprised of a spatial component that uses convolutional filters to capture information about objects and their relationships, and a temporal component that uses large 1D convolutional filters to capture information about how object relationships change across time.
These features are used in tandem with a semi-Markov model that models transitions from one action to another. 
We introduce an efficient constrained segmental inference algorithm for this model 
that is orders of magnitude faster than the current approach. We highlight the effectiveness of our Segmental Spatiotemporal CNN on cooking and surgical action datasets for which we observe substantially improved performance relative to recent baseline methods.

%In the first part of this paper we develop a spatiotemporal model that takes inspiration from early work in robot task modeling by learning how the state of the world changes over time. The spatial component models objects, locations, and object relationships and the temporal component 
%In the second part of the paper we introduce an efficient algorithm for segmental inference that jointly segments and predicts all actions within a video. 

\end{abstract}
\section{Introduction}
\label{sec:intro}
%there has been significant progress in large-scale action classification
New spatiotemporal feature representations \cite{wang_iccv_2013,sun_iccv_2015} and massive datasets like ActivityNet \cite{heilbron_cvpr_2015} have catalyzed progress towards large-scale action recognition in recent years. 
In large-scale action recognition, the goal is to classify diverse actions like skiing and basketball, so it is often advantageous to capture contextual cues like the background appearance. 
In sharp contrast, in fine-grained action recognition, background appearance cues are insufficient to capture the nuances of a complex action, such as subtle changes in object location or state. As a consequence, progress on fine-grained action recognition has been comparatively modest despite active recent developments (e.g. \cite{rohrbach_ijcv_2015,fathi_cvpr_2013,li_cvpr_2015,cheron_iccv_2015,ni_cvpr_2014}).

\begin{figure}[t]
	\center
	\includegraphics[width=.75\hsize]{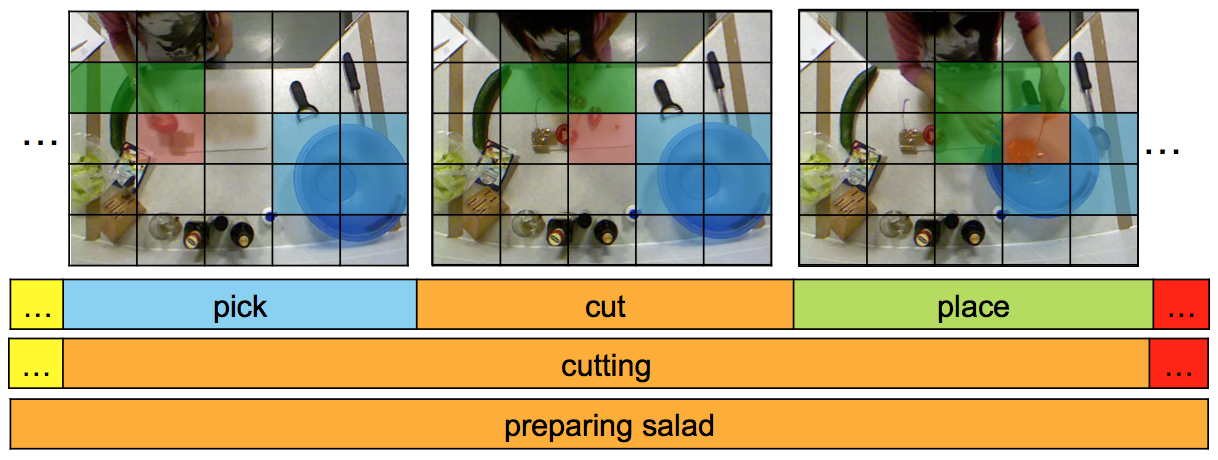}
	\caption{Our model captures object relationships and how these relationships change temporally. (top) Latent \textit{hand} and \textit{tomato} regions are highlighted in different colors on images from the 50 Salads dataset. (bottom) We evaluate on multiple label granularities that model fine-grained or coarse-grained actions.}
	\label{fig:cover}
\end{figure}

In this paper we propose a new approach to fine-grained action recognition that aims to capture information about object states, their relationships, and how they change over time.
Our goal is to temporally segment a video and to classify each of its constituent actions. 
We target goal-driven activities performed in a situated environment, like a kitchen, where a static camera captures a user who performs dozens of actions. % necessary for preparing a salad. 
%where each video is composed of a series of contiguous actions.
For concreteness, refer to the sub-sequence depicted in Figure~\ref{fig:cover}: A user places a tomato onto a cutting board, cuts it with a knife, and places it into a salad bowl. This is part of a much longer salad preparation sequence.
There are many applications of this task including in
industrial manufacturing~\cite{vo_cvpr_2014,hawkins_icra_2014}, surgical training~\cite{JIGSAWS,zappella_media_2013,tao_miccai_2013}, and general human activity analysis (e.g. cooking, sports)~\cite{rohrbach_ijcv_2015,li_cvpr_2015,lei_ubicomp_2012,morariu_cvpr_2011,koppula_ijrr_2013,van_jaise_2010}. 

To address this problem, we introduce a Spatiotemporal Convolutional Neural Network (ST-CNN), which encodes low- and mid- level visual information, and a semi-Markov model that models high-level temporal information. 
%comprised of three components. 
The spatial component of the ST-CNN is a variation on VGG~\cite{VGG} designed for fine-grained tasks which we empirically find captures information about object locations, their states (e.g. whole tomato vs. diced tomato), and inter-object relationships.  
Our network is smaller than models like VGG~\cite{VGG} and AlexNet~\cite{alexnet} and induces more spatial invariance. 
%This is important for applications where there is limited training data. 
%We find that pre-trained networks like VGG and AlexNet are inadequate for differentiating the small motions performed in our applications.
This model diverges from recent fine-grained models, which typically use holistic approaches to model the scene. 

The temporal component of the ST-CNN captures how object relationships change over the course of an action. In the tomato cutting example the \texttt{cut} action changes the tomato's state from \textit{whole} to \textit{diced} and the \texttt{place} action requires moving the tomato from location \textit{cutting board} to \textit{bowl}. 
%Each action is represented as a linear combination of shared temporal convolutional filters. The probability of an action at any given time is computed using 1D convolutions over the spatial activations. These filters are on the order of 10 seconds long and explicitly capture mid-range motion patterns. 
The ST-CNN applies a set of shared temporal 1D convolutional filters to the output of the spatial component. The temporal filters are on the order of 10 seconds long and explicitly capture mid-range motion patterns. The output of these filters is then fed to a linear classifier that computes an action activation score.
The probability of an action at any given time is then estimated by applying a softmax to the action activation scores.

%In contrast to temporal models like Long Short Term Memory (LSTM), which capture mid-range patterns solely using through gating mechanisms, our method achieves superior performance and is more easily interpretable.

The segmental component jointly segments and classifies actions using a semi-Markov model that models pairwise transitions between action segments. 
%Conditional Random Field~\cite{sarawagi_nips_2004}
%captures action-action transitions
%Our inter-action component jointly segment and infer actions using a segmental approach (e.g. ~\cite{sarawagi_nips_2004,pirsiavash_cvpr_2014,shi_ijcv_2011,tao_miccai_2013}).
%van_jaise_2010, tang_cvpr_2012, hu_icra_2014,  
%using a semi-Markov energy model.
%A common approach for performing joint segmentation and classification is to use a chain Conditional Random Field (CRF). 
This model offers two benefits over traditional time series models like linear chain Conditional Random Fields (CRFs) and Recurrent Neural Networks (RNNs). 
Features are computed segment-wise, as opposed to per-frame, and the action at each segment is conditioned on the previous segment instead of the previous frame.
Typically, inference in semi-Markov models is of much higher computational complexity than their frame-wise alternatives. 
In this work we introduce a new constrained inference algorithm that is one to three orders of magnitude faster than the common semi-Markov inference technique.

Despite a large number of action recognition datasets in the computer vision community, few are sufficient for modeling fine-grained segmentation and classification. 
%Common datasets contain too few users or an insufficient amount of data for learning complex models. For example, Georgia Tech Egocentric Activities (GTEA)~\cite{farhadi_cvpr_2009} has seven activities but only four videos per activity. 
%%e are designed for segmentation and classification but have very few users and do not have enough data to train our models. Other datasets, like 
%MPII Cooking~\cite{rohrbach_ijcv_2015} has a larger number of videos but some actions are rarely performed which results in an insufficient ratio of instances per action. 
We apply our approach to two datasets: University of Dundee 50 Salads~\cite{stein_ubicomp_2013}, which is in the cooking domain, and 
the JHU-ISI Surgical Assessment Working Set (JIGSAWS)~\cite{JIGSAWS}, which is in the surgical robotics domain.
Both of these datasets have reasonable amounts of data, interesting task granularity, and realistic task variability.
On these datasets, our model substantially outperforms 
popular methods such as Dense Trajectories, spatial CNNs, and RNNs with Long Short Term Memory (LSTM).

%In summary, in this paper we present a novel model for fine-grained action recognition that
%
%1) Substantially outperforms the state of the art on challenging datasets;
%2) Does so while increasing computational efficiency by one or more orders of magnitude; and
%3) Therefore provides a new methodology for attaching fine-grained action recognition problems.

% which each have a larger number of instances per action.
%We use the University of Dundee 50 Salads~\cite{stein_ubicomp_2013} dataset which includes 50 instances of users preparing salads. There is substantial variability in the order each user performs each action, which ingredients are used, and other stylistic variations. 
%In contrast with many other datasets from the computer vision community, there are many instances of each action. 
%We believe this dataset is useful for addressing many questions in fine-grained action recognition. 
%In addition we validate on the  dataset which is used for surgical skill assessment of robotic training tasks. 

%and show our video-based model achieves superior performance than the state of the art, even compared to some approaches using domain-specific sensor data like the pose of a robot. 
%The insights we describe have implications beyond our problem including on action classification, localization, and detection.
In summary, our contributions are:
\begin{itemize}
\item We develop a Spatiotemporal CNN which we empirically find captures information about object relationships and how relationships change over time,
%\item \TODO{integrate a Semi-markov CRF with a statio-temporal CNNs,}
\item We introduce an efficient algorithm for segmental inference that is one to three orders of magnitude faster than the common approach,
\item We substantially outperform recent methods for fine-grained recognition on two challenging datasets.
%\item We highlight 50 Salads as an important dataset for fine-grained action recognition and provide baselines using Dense Trajectories, VGG, and our model.
\end{itemize}

%The code for our models and the learned features will be released after acceptance of this paper.

% Furthermore, a single label could be used for cutting or mixing any ingredient. Of course to some extent the granularity is application specific. However, perhaps in order to adequately capture a higher level cutting action we need latent states for each action primitive. In this work we investigate the importance of it is to have In this case having finer grained labels may be beneficial.

%Our problem is very difficult for many reasons. To contrast, most action recognition papers classify individual video clips  (typically with short duration) with a single action label. We classify long videos with many sequential action labels. In large scale recognition the labels tend to be very disparate. For example UCF 101 dataset~\cite{UCF} includes \textit{Applying Makeup} and \textit{Archery}. In our task the labels are very similar such as \textit{cutting a cucumber} and \textit{cutting a tomato}. 

%We learn a set of composable action primitives represented by a spatial convolutional filers. We then use our activations as input into a Segmental model. 

\section{Related Work}
\label{sec:related}
%Recent work on activity analysis has focused on tasks like action classification, localization, detection, and segmentation. 
%\footnote{Note that despite our observations highlighting low results, we are not denigrating the impact this research has. We are merely pointing out that these methods are insufficient for practical applications of action recognition.}
\fakesubsection{Holistic Features}
%\textbf{Holistic Features:}
Holistic methods using spatiotemporal features with a bag of words representation are standard for large-scale~\cite{wang_iccv_2013,wang_cvpr_2011,pirsiavash_cvpr_2014,jain_cvpr_2015} and fine-grained \cite{rohrbach_ijcv_2015,fathi_cvpr_2013,li_cvpr_2015,pirsiavash_cvpr_2014,pishchulin_gcpr_2014} action analysis. 
The typical baseline represents a given clip using Improved Dense Trajectories (IDT)~\cite{wang_iccv_2013} 
with a histogram of dictionary elements~\cite{rohrbach_ijcv_2015} or a Fisher Vector encoding~\cite{wang_iccv_2013}. 
Dense Trajectories concatenate HOG, HOF, and MBH texture descriptors extracted along optical flow trajectories to characterize small spatiotemporal patches. 
Empirically they perform well on large-scale tasks, in part because of their ability to capture background detail (e.g. sport arena versus mountaintop). However, for fine-grained tasks the image background is often constant so it is more important to model objects and their relationships. These are typically not modeled in holistic approaches.
%Spatial relationships are typically lost with holistic approaches due to the bag of words or Fisher Vector encoding. 
Furthermore, the typical image patch size for IDT (neighborhood=32px, cell size=2px) is too small to extract high-level object information.

\fakesubsection{Large-scale Action Classification}
Recent efforts to extend CNN models to video~\cite{sun_iccv_2015,jain_cvpr_2015,karpathy_cvpr_2014,simonyan_nips_2014,tran_iccv_2015,peng_thumos_2015,ng_cvpr_2015} improve over holistic methods by encoding spatial and temporal information. However, results from these models are often only superior when CNN features are concatenated with IDT features~\cite{jain_cvpr_2015,tran_iccv_2015,peng_thumos_2015}.
%While recent work has extended CNN models to video~\cite{sun_iccv_2015,jain_cvpr_2015,karpathy_cvpr_2014,simonyan_nips_2014,tran_iccv_2015,peng_thumos_2015,ng_cvpr_2015}, often results are only superior when concatenated with IDT features~\cite{jain_cvpr_2015,tran_iccv_2015,peng_thumos_2015}.
%These CNN models improve over holistic methods by encoding spatial and temporal relationships within an image. 
Furthermore, performance using some of these spatiotemporal CNNs (e.g.~\cite{sun_iccv_2015,karpathy_cvpr_2014,ng_cvpr_2015}) is only marginally better than their spatial-only counterparts or the IDT baselines.
%Yeung \etal~\cite{yeung_cvpr_2016} show improved performance on ActivityNet using less than 2\% of the frames in a video. 
%Several papers (e.g.~\cite{sun_iccv_2015,karpathy_cvpr_2014,ng_cvpr_2015})  have proposed models to fuse spatial and temporal techniques, but results are only marginally better than . 
%Karpathy \etal~\cite{karpathy_cvpr_2014} and Ng \etal~\cite{ng_cvpr_2015} explore various fusion techniques for classifying large-scale actions by combining spatial and temporal information. 
%While each achieve state of the art, their results are only marginally better than IDT baselines. 
%Sun \etal~\cite{sun_iccv_2015} introduced a factorized spatiotemporal CNN to jointly learn spatial and temporal filters. 
%Their temporal component uses 2D temporal convolutions with small filters (e.g. $3\times 3$). 
Our approach is similar in that we propose a spatiotemporal CNN, but our temporal filters are applied in 1D and are much longer in duration.
%This is similar to our first contribution but is targeted at action classification and does not model actions segmentally. 
%Zisserman \etal~\cite{simonyan_nips_2014} introduced a two stream CNN using color and optical flow images for action classification that they show learns features similar to Dense Trajectories. 
%Often CNNs perform worse than IDT unless the two are explicitly combined~\cite{jain_cvpr_2015,tran_iccv_2015,peng_thumos_2015}.

\fakesubsection{From Large-scale Detection to Fine-grained Segmentation}
%Our fine-grained task is more similar to large-scale action detection or localization than classification.
Despite success in classification, large-scale approaches are inadequate for tasks like action localization and detection which are more similar to fine-grained segmentation. In the 2015 THUMOS large-scale action recognition challenge\footnote{THUMOS Challenge: \url{http://www.thumos.info/}}, the top team fused IDT and CNN approaches to achieve 70\% mAP on classification. However, the top method only achieved 18\% (overlap $\geq$ 0.5) for localization. 
Heilbron \etal~\cite{heilbron_cvpr_2015} found similar results on ActivityNet with 11.9\% (overlap $\geq$ 0.2). 
This suggests that important methodological changes are necessary for identifying and localizing actions regardless of fine-grained or large-scale.  

Moving to fine-grained recognition, recent work has combined holistic methods with human pose or object detection.
On MPII Cooking, Rohrbach \etal~\cite{rohrbach_ijcv_2015} combine IDT with pose features to get a detection score of 34.5\% compared to 29.5\% without pose features.
Cheron \etal~\cite{cheron_iccv_2015} show that if temporal segmentation on MPII is known then CNN-based pose features achieve 71.4\% mAP. While this performance is comparatively high, classification is a much easier problem than detection. 
%Cheron \etal \cite{cheron_iccv_2015} show that if classification is known then it is possible to achieve 71.4\% mAP on MPII Cooking using human pose features, however, they do not apply it to detection. 
%Rohrbach \etal~\cite{rohrbach_ijcv_2015} found similar results on the MPII Cooking dataset where they achieve a detection score of 29.5\% mAP with an IDT baseline.
Object-centric methods (e.g.~\cite{fathi_cvpr_2013,li_cvpr_2015,ni_cvpr_2014}), first detect the identity and location of objects in an image. 
Ni \etal~\cite{ni_cvpr_2014} achieve 54.3\% mAP on MPII Cooking and 79\% on the ICPR 2012 Kitchen Scene Context-based Gesture Recognition dataset.
While performance is state of the art, their method requires learning object models from a large number of manual annotations. 
%and may not scale to new environments. 
In our work we learn a latent object representation without object annotations.
Lastly, on Georgia Tech Egocentric Activities, Li \etal~\cite{li_cvpr_2015}
use object, egocentric, and hand features to achieve 66.8\% accuracy for action classification versus an IDT baseline of 39.8\%. 
Their features are similar to IDT but they use a recent hand-detection method to find the regions of most importance in each image. 
%In contrast to our approach, their method does not appear to capture the relationship between objects.

%This is a admirable but may not scale well to new kitchens. 
%In 50 Salads there are many irrelevant objects in view at all times so an action cannot be sufficient recognized based on objects in the scene. \TODO{check details from paper}
%Furthermore, if we were to scale our approach to new kitchens, we are not guaranteed to have a model of each object.  
%(In addition they use course and fine granularity actions for prediction).
%Yezhou \etal \cite{yang_cvpr_2013} learn how objects change as a function of an action. For example, in a cutting task the transition from a whole cucumber to two halves is modeled as a change from one segment to two. It is not clear that this can easily be applied in settings like ours with substantial clutter and occlusion. 
%Additional approaches using bag of words models have achieved moderate success for classification \cite{mpii} in the case where temporal segmentation is known. This is special case of the joint segmentation and classification problem. For completeness we also evaluate our models for this problem. 

\fakesubsection{Temporal Models}
Several papers have used Conditional Random Fields for action segmentation and classification (e.g. ~\cite{tao_miccai_2013,pirsiavash_cvpr_2014,shi_ijcv_2011,tang_cvpr_2012}). 
CRFs offer a principled approach for combining multiple energy terms like segment-wise unaries and pairwise action transitions. 
Most of these approaches have been applied to simpler activities like recognizing \textit{walking} versus \textit{bending} versus \textit{drawing}~\cite{shi_ijcv_2011}.
In each of the listed cases, segments are modeled with histograms of holistic features. In our work segments are modeled using spatiotemporal CNN activations.
%For example Pirsiavash and Ramanan~\cite{pirsiavash_cvpr_2014} propose a segmental regular grammar for decomposing actions into latent action primitives for sports videos using IDT. 
%This model can be viewed as a latent Semi Markov CRF if the objective is used as the energy in a Gibbs distribution.

Recently, there has been significant interest in RNNs, specifically those using LSTM (e.g.~\cite{ng_cvpr_2015,Vinyals_2015_CVPR,lrcn2014}). RNNs with LSTM use gating mechanisms to implicitly learn how latent states transition within and between actions.
While their performance is often impressive, they are black box models that are hard to interpret.
In contrast, the temporal component of our CNN explicitly learns how latent states transition and is easy to interpret and visualize. It is more similar to models in speech recognition (e.g.~\cite{ng_2015,abdel_interspeech_2013}), which learn phonemes using 1D convolutional filters, or in robotics, which learn sensor-based action primitives~\cite{lea_icra_2016}.
For completeness we compare our model with LSTM-based RNNs.

\section{Spatiotemporal CNN Model}
\label{sec:model}

\begin{figure}[t]
%\center
%	\begin{table}
		\begin{minipage}[T]{0.5\linewidth}
			\begin{center}			
%							Full Model\\
			\includegraphics[width=.99\hsize]{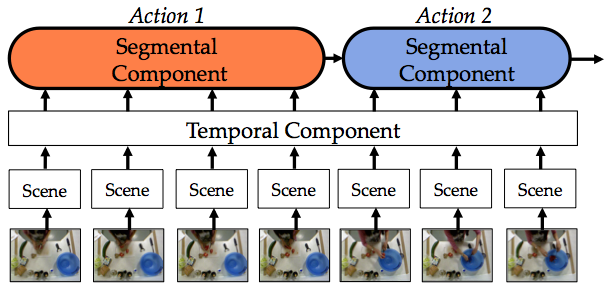}\\
			Full Model\\
	 		\end{center}			
		\end{minipage}
		\begin{minipage}[T]{0.5\linewidth}
			\begin{center}			
%			\vspace{25pt}
			\includegraphics[width=.99\hsize]{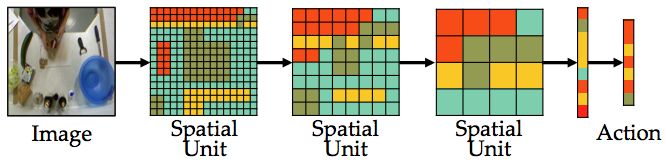}\\
			  Spatial Model\\
			  \end{center}
		\end{minipage}
%	\end{table}
%	\includegraphics[width=.49\hsize]{imgs/ST-CNN}
	\caption{(left) Our model contains three components. The spatial, temporal, and segmental units encode object relationships, how those relationships change, and how actions transition from one to another. (right) The spatial component of our model.}
	\label{fig:model}
\end{figure}

In this section we introduce the spatial and temporal components of our ST-CNN. 
The input is a video including a color image and a motion image for each frame. The output is a vector of action probabilities at every frame.
Figure~\ref{fig:model} (left) depicts the full Segmental Spatiotemporal model.
%In this section we describe the spatial and temporal components and in Section~\ref{sec:inference} describe segmental inference.

\subsection{Spatial Component}

%In recent years CNNs have become commonplace. Models like VGG \cite{VGG} and Alexnet \cite{Alexnet} have been wildly successful for many problems. 
In this section, we introduce a CNN topology inspired by VGG~\cite{VGG} that uses hierarchical convolutional filters to capture object texture and spatial location. First we introduce the mathematical framework, as depicted in Figure~\ref{fig:model} (right), and then highlight differences between our approach and other CNNs. For a recent introduction to CNNs see~\cite{Goodfellow_book}.
%related models like VGG and AlexNet \cite{alexnet}. 
% depicts the spatial model.

%is modeled as follows. First, a set of spatial units convolve an image with a set of filters, the pixels are pooled locally to retain the best feature responses, the features are concatenated into a large feature vector. Then a set of fully connected layers learn which feature responses correspond to which classes. In this section we describe how this idea can naturally be used to model how objects interact with each other for fine-grained recognition tasks.

%We do not claim that our spatial model is novel, however, our interpretation for fine-grained action recognition is unique and its use with the temporal component in this context is new.

For each time $t$ there is an image pair $I_t = \{I_t^c, I_t^m\}$, where $I_t^c$ is a color image and $I_t^m$ is a Motion History Image~\cite{davis_cvpr_1997}. The motion image captures when an object has moved into or out of a region and is computed by taking the difference between frames across a 2 second window.
Other work (e.g.~\cite{simonyan_nips_2014}) has shown success using optical flow as a motion image. We found optical flow to be insufficient for capturing small hand motions and noisy due to the video compression.

The image pair $I_t$ is fed into a CNN with $L$ spatial units, each of which is composed of a convolutional layer with $F_l$ filters of size $3 \times 3$, a Rectified Linear Unit (ReLU), and $3 \times 3$ max pooling. The output of each unit is $r^l = \{r^l_i\}_{i=1}^{R_l}$, where $r_i^l \in \mathbb{R}^{F_l}$, is the activation vector for a specific region in an image. For an $N_l \times N_l$ grid there are $R_l = N_l^2$ regions as depicted by the colored blocks in Figure~\ref{fig:model} (right).

\begin{figure*}[t]
\center
	\includegraphics[width=.9\hsize]{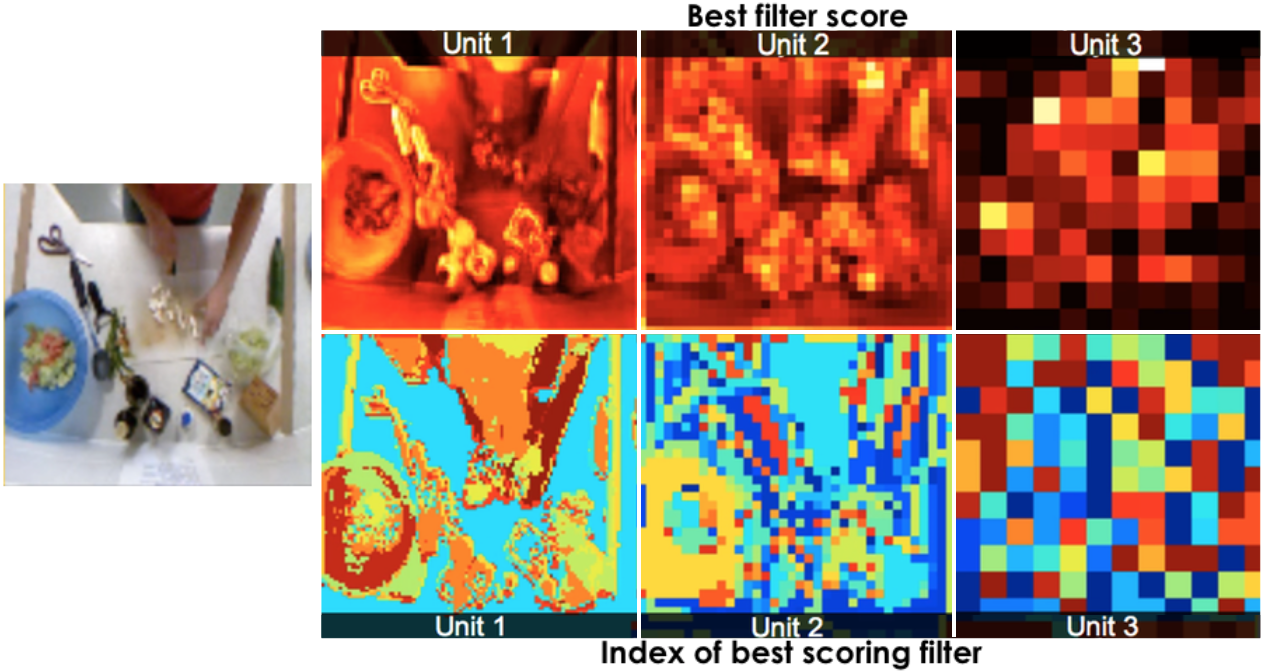}
	\caption{The user is chopping vegetables. The top images show the best filter activations after each convolutional unit from the CNN. The activations around the cutting board and bowl are high (yellow)  whereas in unimportant regions are low (black/red). 
	The bottom images indicate which filter gave the highest activation for each region. Each color corresponds to a different filter index. }	
	\label{fig:activations}
\end{figure*}

The output of the $L$ spatial units is fed into a fully connected layer which has $F_{fc}$ states that capture relationships between regions and their corresponding latent object representations.
For example, a state may produce a high score for \textit{tomato} in the region with the cutting board and \textit{knife} in the region next to it. 
The state $h \in \mathbb{R}^{F_{fc}}$ is a function of weights $W^{(0)} \in \mathbb{R}^{F_{fc} \times F_L R_L}$ and biases $b^{(0)} \in \mathbb{R}^{F_{fc} }$ where $r^L \in \mathbb{R}^{F_L R_L }$ is the concatenation of activations in all regions after the $L$th spatial unit:\footnote{For notational clarity we denote all weight matrices as $W^{(\cdot)}$ and bias vectors $b^{(\cdot)}$ to reduce the number of variables.}
\begin{align}
h = \text{ReLU}(W^{(0)} r^L + b^{(0)}).
\end{align}

The above spatial component, when applied to frame $t$, produces state vector $h_t$.
Ideally, the spatial and temporal components of our CNN should be trained jointly, but this requires an exorbitant amount of GPU memory. Therefore, we first train the spatial model and then train the temporal model.
As such, we train the spatial component with auxiliary labels, $z$. Let $z_t \in \{0,1\}^C$, where C is the number of classes, be the ground truth action label for each time step. We predict the probability, $\hat{z}_t \in [0,1]^C$, of each action class at that frame using the softmax function: 
\begin{align}
\hat{z}_t = \text{softmax}(W^{(1)} h_t + b^{(1)}).
\end{align}
where $W^{(1)} \in \mathbb{R}^{C \times F_{f_c}}$ and $b^{(1)} \in \mathbb{R}^C$.
Note, $\hat{z}_t$ is computed solely for the purpose of training the spatial component. The input to the temporal component is  $h_t$.

Figure~\ref{fig:activations} shows example CNN activations after each spatial unit. The top row shows the sum of all filter activations after that layer and the bottom row shows the color corresponding to the best scoring filter at that location. We find that these filters are similar to mid-level object detectors. Notice the relevant objects in the image and the regions corresponding to the action all have high activations and different best-scoring filters.
%This We observe that we automatically learn a 
%While it may appear advantageous to learn a set of detectors for the relevant objects in our scene, we find that our model automatically learns mid-level object detectors purely from the action annotations. See feature responses in Figure~\ref{fig:activations} for examples.

%We don't want to learn a model for each individual object so we learn a latent represention of where objects are in the scene and their state. In particular we break the image into different regions which will encode the salient object in that area. 

%In our applications the absolute and relative locations of objects is important. A common way to approach this is to create a grid CRF that models each region using a unary and pairwise term. However, we would like to capture relationships between regions that are not adjacent, such as between the cutting board and salad bowl. However, the computional complexity of using a higher-order CRF grows exponentially. We choose to learn a set of hidden states connecting all regions in the hopes that individual nodes will learn associations with specific regions. 

\fakesubsection{Relationships to other CNNs}
Our network is inspired by models like VGG and AlexNet but differs in important ways. Like VGG, we employ a sequence of spatial units with common parameters like filter size. However, we found that using two consecutive convolution layers in each spatial unit has negligible impact on performance. 
Normalization layers, like in AlexNet, did not improve performance either. 
Overall our network is shallower, has fewer spatial regions, and contains only one fully connected layer.
In addition, common data-augmentation techniques, including image rotation and translation, introduce unwanted spatial and rotational invariances, which have a negative impact on our performance. 
% becuase they . These augmentations destroy becuase they induce 
%The biggest impact on our performance is the number of spatial units and size of our regions. 

We performed cross validation using one to seven spatial units and grid sizes from $1 \times 1$ to $9 \times 9$ and found three spatial units with a $3 \times 3$ grid achieved the best results. 
By contrast, for image classification, deep networks tend to use at least four spatial units and have larger grid counts. VGG uses a $7 \times 7$ grid and AlexNet uses a $12 \times 12$ grid. 
%Our findings may be due to the increased spatial invariance in the model due to having fewer grid elements. 
A low spatial resolution naturally induces more spatial invariance, which is useful when there is limited amounts of training data.
%may be better for applications like ours where 
% fine-grained tasks because its larger grid elements naturally induce more spatial invariance. 
%Furthermore, a smaller resolution grid requires 
To contrast, if the grid resolution is larger, more training data is necessary to capture all object configurations.
% which is important for fine-grained applications when training data is often limited.
We compare the performance of our model with a pre-trained VGG network in the results.
%where there is a limited amount of training data and the precise location of each object is less important than the relative locations.
%In our cooking application, knowing that the hands and cutting board are in the same region matters more than knowing they are in a specific region.

%We also experimented using pre-trained VGG and AlexNet models with limited success. These networks tend to be good at distinguishing between vastly different objects or scenes but not at distinguishing minor changes within one scene. 

%A higher resolution grid requires more training data model because larger grid elements induce more spatial invariance.
%the precise location of the bowl is not important; it is typically somewhere in the lower third of the image. It does not matter if it is at the very bottom or a quarter of the way up. 
% which is important if there are a limited number of samples. 
%In the cooking application, the salad bowl may be located in a different element within each video. If the grid resolution is high then we need more videos to capture its location whereas if the resolution is low then the bowl will be located in the same element in more videos. 

%-------------------------------------------------------- 
%-------------------------------------------------------- 
\subsection{Temporal Component}

%The spatial component is only sufficient for modeling objects, their locations, and states. 
Temporal convolutional filters capture how the scene changes over the course of an action. These filters capture properties like the scene configuration at the beginning or end of an action and different ways users perform the same action. 

%We assume that each action can be classified with a linear combination of shared temporal filters.
%, like those shown in  Figure~\ref{fig:weights}. 
For video duration $T$, let $h = \{h_t\}_{t=1}^T$ be the set of spatial features and $y_t \in \{1,...,C\}$ be an action label at time $t$.
%For convenience we define $H_{t:t+d}$ to be a sequence of features from time $t$ to $t+d-1$. 
%Let the collection of features $x_t$ for $t \in \{1,\dots, T\}$ from the spatial component be $X \in \mathbb{R}^{C \times T}$ and 
We learn $F_e$ temporal filters $W^{(2)} = \{W^{(2)}_1, \dots, W^{(2)}_{F_e}\}$ with biases $b^{(2)}= \{ b^{(2)}_1, \dots, b^{(2)}_{F_e} \}$ shared across actions. 
Each filter is of duration $d$ such that $W^{(2)}_i \in \mathbb{R}^{ d \times F_{fc}}$.
%\footnote{Here we denote the filters and segments as matrices for clarity. In practice everything is vectorized.} 
The activation for the $i$-th filter at time $t$ is given by a 1D convolution between the spatial features $h$ and the temporal filters using a ReLU non-linearity:
%$H_{t:t+d}$
\begin{align}
%\text{ReLU}(W_i^{(2)} \ast H_{t:t+d} + b_i^{(2)}) =
	a_{t,i} =  \text{ReLU}(\sum_{t'=1}^{d} W_{i,t'}^{(2)} h_{t+d-t'} + b_i^{(2)}).
\end{align}

A score vector $s_t \in \mathbb{R}^C$ is a function of weight vectors $W^{(3)} \in \mathbb{R}^{C \times F_e}$ and biases $b^{(3)} \in \mathbb{R}^{C}$ with the softmax function:
\begin{align}
	s_t =  \text{softmax}(W^{(3)}  a_t + b^{(3)}).
\end{align}
%In Section~\ref{sef:inference} we describe inference in our model using scores $\{s_t\}_{t=1}^T$. 
%We index the $c$-th class using $s_t^c$.

We choose filter lengths spanning 10 seconds of video. This is much larger than in related work (e.g.~\cite{sun_iccv_2015,ng_cvpr_2015}). Qualitatively, we found these filters capture states, transitions between states, and attributes like action duration. 
In principle, we could create a deep temporal model. Multiple layers did not improve performance in preliminary experiments, however, it is worth further exploration.

%\begin{figure}
%\center
%	\includegraphics[width=.99\hsize]{imgs/spatial.png}
%	\caption{\TODO{COMBINE WITH COVER}}
%	\label{fig:spatial}
%\end{figure}

%\begin{figure}
%\center
%	\includegraphics[width=.99\hsize]{imgs/primitives.png}
%	\caption{DON'T ACTUALLY USE THIS IMAGE.... BUT MAYBE SOMETHING LIKE THIS?}
%	\label{fig:temporal}
%\end{figure}

%\THOUGHT{Interpretability}

\subsection{Learning}
%While it may be advantageous to learn all spatial and temporal parameters jointly,
%it makes it more difficult to perform ablative and comparative analysis. \TODO{technically could... but requires a ton of memory or slicing up the videos}
%Thus we choose to learn the scene and temporal models independently. 
We learn parameters $W=\{W^{0}, W^{1}, W^{2}, W^{3}\}$, $b=\{b^{0}, b^{1}, b^{2}, b^{3}\}$, and the convolutional filters with the cross entropy loss function.
%Let $Y \in \mathbb{R}^{C \times T}$ be a matrix where the true class at each time step is 1 and the rest are 0.
We minimize the spatial and temporal network losses independently
%$e(Y, X) = \| Y - X \|_F^2$
%$e(Y, S) = \| Y - S \|_F^2$ for the temporal network
%Recall that $X$ is the set of spatial features and $S$ is the set of scores for each time step. 
using ADAM~\cite{ADAM}. 
%, a recent method for stochastic optimization
Dropout regularization is used on fully connected layers.
Parameters such as grid size, number of filters, and non-linearity functions are determined from cross validation using one split each from the datasets described later. We use $F=\{ 64, 96, 128 \}$ filters in the three corresponding spatial units and $F_{fc}=256$ fully connected states.
We used Keras~\cite{keras}, a library of deep learning tools, to implement our model.

%Each region is encoded with feature vector $r_i \in \mathbb{R}^{F_I}$
% for $i \in \{1, \dots, R\}$ 
%which is computed as a set of convolutions over that region. 

%. On top of the spatial units is one fully connected layer with 256 nodes (using sigmoid activation) and one fully connected layer with $C$ nodes where $C$ is the number of action classes. Dropout is used on the fully connected layers. 
%\begin{align}
%%r_i = [REDO MATH]
%r_i = f(w^{(0)} \ast I^{R_i}_t + )
%\end{align}

\section{Segmental Model}
\label{sec:inference}
%In this section we define our segmental model and introduce an efficient algorithm for performing inference. Later we show that this model improves performance by reducing the number of false-positive action predictions. 
%Note that this is in contrast to most models, including many CNNs, RNNs, and CRFs, which predict the label individually for each frame: $y_t = \argmax\limits_{y} s_t^y$. 

%\fakesubsection{Segmental Model}
%Action-to-action transitions and temporal priors are modeled 
We jointly segment and classify actions with a variation on a semi-Markov model that uses activations from the ST-CNN and a pairwise term to capture segment-wise action-to-action transitions.
Similar models have been proposed by Sarawagi and Cohen~\cite{sarawagi_nips_2004} for Semi-Markov CRFs, by Shi \etal~\cite{shi_ijcv_2011} for discriminative semi-Markov models and by Pirsiavash and Ramanan \cite{pirsiavash_cvpr_2014} for Segmental Regular Grammars.
We show a more efficient inference algorithm made possible by re-formulating the traditional segmental inference problem. 

We start with notation equivalent to Sarawagi and Cohen~\cite{sarawagi_nips_2004}. Let tuple $P_j=(y_j, t_j, d_j)$ be the $j$th action segment, where $y_j$ is the action label, $t_j$ is the start time, and $d_j$ is the segment duration. There is a sequence of $M$ segments $P=\{P_1,...,P_M\}$, $0 < M \leq T$, such that the start of segment $j$ coincides with the end of the previous segment, i.e. $t_j=t_{j-1}+d_{j-1}$, and the durations sum to the total time $\sum_{i=1}^M d_i = T$.
Given scores $S=\{s_1, \dots, s_T\}$, we infer segments $P$ that maximize total score $E(S,P)$ for the video using segment function $f(\cdot)$:
%, and temporal prior $p(\cdot)$:
%transition matrix $A \in \mathcal{R}^{C \times C}$, and temporal prior $P \in \mathcal{R}^{C \times M}$:
%\begin{align}
%	E(S,P)=\sum_{j=1}^M f(S, y_j, t_j, d_j) + A_{y_{j-1},y_j} + P^j_{y_j}
%\end{align}
\begin{align}
%	E(S,P)=\sum_{j=1}^M f(S, y_j, t_j, d_j) + A_{y_{j-1},y_j}
	E(S,P)=\sum_{j=1}^M f(S, y_{j-1}, y_j, t_j, d_j).
%	E(S,P)=\sum_{j=1}^M f(S, y_j, t_j, d_j) + p(y_j, t_j, d_j) + A_{y_{j-1},y_j}	
\end{align}
%This model is a Conditional Random Field where $\text{Pr}(P|S) \propto \exp(E(S,P))$.
Our segment function is the sum of ST-CNN scores 
plus the transition score for going from action $y_{j-1}$ at segment $j-1$ to action $y_j$ at segment $j$:
%contains transition matrix $A \in \mathbb{R}^{C \times C}$ and :
%We impose the constraint $y_i \neq y_{i-1}$ for all segments $i$ to avoid the degenerate solution where each consecutive segment receives the same label (e.g. $y_1=y_2=...=a$). We choose $f$ to be the sum function with a constraint on self-transitions:
\begin{align}
%f(S, y_j, t_j, d_j) = \frac{1}{d_j} \sum_{t=t_j}^{t_j+d_j-1} S^{y_j}_t
f(S, y_{j-1}, y_j, t_j, d_j) = A_{y_{j-1},y_j} +  \sum_{t=t_j}^{t_j+d_j-1} s_{t,y_j}
%\frac{1}{d_j}
\end{align}
%
%$P^j_{y_j}$ is the log probability of segment $j$ belonging to class $y_j$. For the temporal prior we resample each sequence to be of the same length $T'$. The prior for segment $j$ to belong to class $c$ is given by:
%\begin{align}
%P^j_c = \frac{1}{t'_{j+1}-t'_{j}} \sum_{t=t'_j}^{t'_{j+1}} \mathbf{1}[y_t=c]
%\end{align}
%

%The temporal prior models the likelihood of a given action happening at a given time and is parameterized by matrix $B \in \mathbb{R}^{C \times T'}$ where $T'$ is a canonical sequence length. Each parameter in $B^t_{c}$ is computed by resampling each sequence to be of length $T'$ and computing the per-class probability for each timestep $t$ and class $c$.
%Notably, sequences tend to start in the same action and end in the same actions. 
%$P^j_{y_j}$ is the log probability of segment $j$ belonging to class $y_j$. For the temporal prior we resample each sequence to be of the same length $T'$. 
%The prior for segment $j$ to belong to class $c$ is given by: 
%The prior is given by: 
%\begin{align}
%p(y_j, t_j, d_j) = \frac{1}{d_j} \sum_{t=t_j}^{t_j+d_j-1} B^t_{y_j}
%\end{align}
%%
%\TODO{One benefit of having a prior function, as opposed to a single parameter for each segment $j$, is that it better models }
The entries of $A \in \mathbb{R}^{C \times C}$ 
%is the score for transitioning from action $y_{j-1}$ at segment $j-1$ to $y_{j}$ at segment $j$. Entries of $A$ are 
are estimated from the training data as the log of the probabilities 
of transitioning from one action to another.

%The segmental transition matrix is defined as the 
%The transition $A_{y_{j-1},y_j}$ is given by the log probability of going from class $y_{j-1}$ to $y_j$ and is learned directly from the training data.

%These parameters are learned by computing the corresponding probability distributions from the the training data.

%\begin{align}
%	f(S, y_j, t_j, d_j) = 
%\begin{cases}
%\sum_{t=t_j}^{t_j+d_j-1} S^{y_j}_t  ,& \text{if } y_j \neq y_{j-1}\\
%    -\infty,              & \text{otherwise}
%\end{cases}	
%\end{align}

%\begin{align}
%f(S, y_j, t_j, d_j) = 
%\begin{cases}
%P^j_{y_j} + T_{y_{j-1},y_j} + \frac{1}{d_j} \sum_{t=t_j}^{t_j+d_j-1} S^{y_j}_t  ,& \text{if } y_j \neq y_{j-1}\\
%-\infty,              & \text{otherwise}
%\end{cases}	
%\end{align}

\subsection{Segmental Inference}
The traditional semi-Markov inference method 
%The inference method first proposed by Sarawagi and Cohen~\cite{sarawagi_nips_2004} for Semi-Markov CRFs, and reintroduced by Shi \etal~\cite{shi_ijcv_2011} for discriminative semi-Markov models and by Pirsiavash and Ramanan \cite{pirsiavash_cvpr_2014} for Segmental Regular Grammars,
solves the following discrete optimization problem 
\begin{align}
 \max\limits_{P \in \mathcal{P}}  \text{ } E(S, P) ~~  s.t.~\textstyle \quad t_j = t_{j-1}+d_{j-1} ~ \forall j ~~  \mbox{and} ~~ \sum_{i=j}^M d_j = T .
%\\ M \in \{1,\dots, T\}
%P_1,\dots, P_M}
\end{align}
%where $P^*$ is the best labeling and 
where $\mathcal{P}$ is the set of all segmentations.
%\begin{align}
%P^* = \argmax\limits_{P \in \mathcal{P}}  \text{ } E(S, P) ~~  s.t.~\textstyle \quad t_j = t_{j-1}+d_{j-1}+1 ~ \forall j ~~  \mbox{and} ~~ \sum_{i=j}^M d_j = T .
%%\\ M \in \{1,\dots, T\}
%%P_1,\dots, P_M}
%\end{align}
%Sarawagi and Cohen introduced an algorithm, 
%We refer to the inference algorithms of~\cite{sarawagi_nips_2004} and~\cite{shi_ijcv_2011} as Segmental Viterbi because they extend the traditional Viterbi method to the semi-Markov case. 
The optimal labeling is typically found using an extension of the Viterbi algorithm to the semi-Markov case, which we refer to as Segmental Viterbi~\cite{sarawagi_nips_2004,shi_ijcv_2011,pirsiavash_cvpr_2014}.
The algorithm recursively computes the score $V_{t,c}$ for the best labeling whose last segment ends at time $t$ and is assigned class $c$:
%for each time $t$ and class $c$ where $t$ corresponds to the ending time for a segment:
\begin{align}
V_{t,c} = \max\limits_{\substack{d \in \{1 \dots D\} \\ c' \in \mathcal{Y} \backslash c }} V_{t-d,c'} +  f(S, c', c, t-d, d).
\end{align}
%The forward pass is shown in Algorithm~\ref{alg:seg_viterbi}. 
The optimal labels are recovered by backtracking through the matrix $V$ using the predicted segment durations.

This approach is inherently frame-wise: for each frame, compute scores for all possible segment durations, current labels, and previous labels. 
%Their recursive update step is defined as follows with score $V_{y,t}$ for class $y$, time $t$, previous class $y'$ and segment duration $d$: 
%$V_{y,t}=\argmax\limits_{d, y'} V_{y',t-d} + f(S, y, t, d)$. 
This results in an algorithm of complexity $O(T^2C^2)$, in the naive case, because the duration of each segment ranges from $1$ to $T$.
If the segment duration is bounded then complexity is reduced to $O(TDC^2)$, where $D$ is the maximum segment duration~\cite{sarawagi_nips_2004,shi_ijcv_2011,pirsiavash_cvpr_2014}. 
%is typically maximized over values from $1$ to the maximum segment length $D$. 
%In the forward pass the scores are computed for all timesteps, classes, previous classes, and durations. 
% The computational complexity of this approach is $O(TC^2D)$.

% in addition to optimizing over the they optimizes over all possible durations of each segment. They perform dynamic programming with score table $V$ using the recursive update 
%$V_{c,t}=\argmax\limits_{\substack{d=1...D \\ y'}} V_{y',t-d} + f(X, c, t, d)$
%$V_{c,t}=\argmax\limits_{d, y'} V_{y',t-d} + f(s, c, t, d)$
% In the forward pass the scores are computed for all timesteps, classes, previous classes, and durations. The optimal labels are recovered by backtracing through the matrix. 

To further accelerate the computation of the optimal labels, 
we introduce an alternative approach in which we constrain the number of segments, $M$, by an upper bound, $K$, such that $0 < M \leq K$. 
If $K=T$, this is equivalent to that of the previous problem. 
Furthermore, we remove the duration variables $d_j$, which are redundant given all times $t_j$, and simplify the segment notation to be $\hat{P_j}=(y_j, t_j)$. 
Now, instead of adding constraints on the durations of each segment, we only require that the start of the $j$th segment comes after segment $j-1$. 
%We no longer optimize over the duration so we remove that from our objective. 
%We now optimize over the segment count instead of over each segment duration. 
%For comparison, Segmental Viterbi allows for $0 < M \leq T$. 
%Instead of optimizing over segment durations, we optimize over the number of segments in a sequence. 
We solve the problem
%\begin{align}
%P^* = \argmax\limits_{P_1,\dots, P_M} \text{ } E(S, P) \quad
%s.t. \textstyle   \quad   0 < M \leq K 
%\end{align}
\begin{align}
%P^* = \argmax\limits_{\substack{\hat{P}_1,\dots, \hat{P}_M \\ M \in \{1,\dots, K\}}} \text{ } E(S, \hat{P}) \quad s.t.~\textstyle \quad t_{j-1} < t_{j} .
%\hat{P}^* = \argmax\limits_{\substack{M \in \{1,\dots, K\} \\ \hat{P} \in \hat{\mathcal{P}}_M}} \text{ } E(S, \hat{P}) \quad s.t.~\textstyle \quad t_{j-1} < t_{j} ~ \forall j \in \{1, \dots, M\}
\max\limits_{\substack{M \in \{1,\dots, K\} \\ \hat{P} \in \hat{\mathcal{P}}_M}} \text{ } E(S, \hat{P}) \quad s.t.~\textstyle  \quad t_{j-1} < t_{j} ~ \forall j \in \{1, \dots, M\}  
%t_0=1 ~ \mbox{and}
\end{align}
where $\hat{\mathcal{P}}_M$ is the set of all segmentations with $M$ segments. 
The $d$ found in function $f(\cdot)$ is now simply $t_j-t_{j-1}$. 
As a byproduct, this formulation prevents gross over-segmentations of the data and thus improves performance. 

We assume the score for each segment is the sum of scores over each frame, so the total score for the segmentation containing $k$ segments can be recursively computed using the scores for a segmentation containing $(k-1)$ segments. 
Specifically, we first compute the best segmentation assuming $M=1$ segments, then compute the best segmentation for $M=2$ segments, up to $M=K$ segments.
%Specifically, we compute all scores assuming $M=1$, then compute scores for $M=2$, up to $M=K$.
%Our algorithm maximizes over the start time, current label, and previous label for each segment. 
%This can be computed efficiently formulating it as a constrained optimization problem where $K$ is an upper bound on the number of segments:
Let $\bar{V}^{k}_{t,c}$ be the score for the best labeling with $k$ segments ending in class $c$ at time $t$:
%\begin{align}
%\bar{V}^{k}_{t,c} = \max\limits_{\substack{d \in \{1 \dots D\} \\ c' \in \mathcal{Y}/c }} \bar{V}^{k'}_{t-d,c'} +  f(S, y_{j-1}, y_j, t_j, d_j)
\begin{align}
%\bar{V}^{k}_{t,c} = \max\limits_{\substack{ c' \in \mathcal{Y} }} \bar{V}^{k'}_{t-1,c'} +  A_{c',c} + s_{t,c}
\bar{V}^{k}_{t,c} = \max \Big(~~\max_{ c ‘ \in \mathcal{Y} \setminus c}  ( \bar{V}_{t-1,c'}^{k-1} + A_{c',c} ), ~~ \bar{V}_{t-1,c}^k ~~\Big) + s_{t,c}.
\end{align}
This recursion contains two cases: (1) if transitioning into a new segment ($c' \neq c$), use the best incoming score from the previous segment $k-1$ at $t-1$ and (2) if staying in the same segment ($c' = c$), use the score from the current segment at $t-1$. 
%where, if staying in the same segment ($c=c'$), then $k'= k$ and $A_{c',c} = 0$, otherwise $k'= k-1$ and $A_{c',c}$ is the pairwise term.
% each new segment is a function of the previous scores $V^{'k-1}_{\cdot,c}$.
%The best starting time for each segment is given by $V^{'k}_{t,c}$ for having $k$ segments in the sequence.
%For each consecutive segment, we reuse $V^{'k}_{t,c}$ to calculate the best score for the $(k+1)$ segment starting at $t$.
%We start by computing the best scores assuming there is only one segment. From there, we can leverage those scores to find the best sequence with two segments. 
%\begin{align}
%V^{'k}_{t,c} = \max\limits_{c' \in \mathcal{Y}} V^{'k-1}_{t-1,c'} + f(S, c', c, t_j, d_j)
%%V^{'k}_{t,c} = \max\limits_{\substack{t_k \in \{k, k-1\} \\ c' \in \mathcal{Y}/c }} V^{'k-1}_{t-1,c'} + f(S, c', c, t_j, d_j)
%%V^{'k}_{t,c} = \max\limits_{c' \in \mathcal{Y}}
%%\begin{cases}
%%V^{'k-1}_{t-d_j,c'} + f(S, c', c, t, d_j) & \text{if } c \neq c'\\
%%V^{'k}_{t-d_j,c} + f(S, c, c, t, )   & \text{otherwise}
%%V^{'k-1}_{t-1,c'}  + A_{c',c}  & \text{if } c \neq c'\\
%%V^{'k}_{t-1,c'}                  & \text{otherwise}
%%\end{cases}
%\end{align}
Our forward pass, in which we compute each score $\bar{V}_{t,c}^k$, is shown in Algorithm~\ref{alg:seg_inference}. The optimal labeling is found by backtracking through $\bar{V}$. 
\begin{center}
\begin{minipage}[t!]{.7\textwidth}
	\begin{algorithm}[H]
		\begin{algorithmic}[0]
			\For{$k=1:K$ } %\Comment{Segments}
			\For{$t=k:T$ } %\Comment{Time steps}		
			\For{$c=1:C$ } %\Comment{Classes}						
			\State $v_{cur} = \bar{V}^{k}_{t-1,c} $
			\State $v_{prev} = \max\limits_{c' \in \mathcal{Y} \backslash c} \bar{V}^{k-1}_{t-1,c'} + A_{c',c}$
			\State $\bar{V}^{k}_{t,c} = \max(v_{cur}, v_{prev})+s_{t,c}$ 
			\EndFor
			\EndFor
			\EndFor
			%		\vspace{6.25pt}
		\end{algorithmic}
		\caption{Our Semi-Markov Forward Pass}
		\label{alg:seg_inference}
	\end{algorithm}
\end{minipage}
\end{center}

The complexity of our algorithm, $O(KTC^2)$, is $\frac{D}{K}$ times more efficient than Segmental Viterbi assuming $K < D$. 
%Clearly it becomes more efficient as the ratio of the number of segments $K$ to the maximum segment duration $D$ decreases. 
In most practical applications, $K$ is much smaller than $D$. In the evaluated datasets there is a speedup of one to three orders of magnitude.
Note, however, our method requires $K$ times more memory than Segmental Viterbi. Ours has space complexity $O(KTC)$, whereas Segmental Viterbi has complexity of $O(TC)$. Typically $K \ll T$ so the increase in memory is easily manageable on any modern computer. 
In all cases, we set $K$ based on the maximum number of segments in the training split.

\section{Experimental Setup}
\label{sec:evaluation}

Historically, most action recognition datasets were developed for classifying individual actions using pre-trimmed clips. 
Recent datasets for fine-grained recognition have been developed to classify many actions, however they often contain too few users or an insufficient amount of data to learn complex models. 
% or are targeted at specific tasks like egocentric video. 
%e are designed for segmentation and classification but have very few users and do not have enough data to train our models. Other datasets, like 
MPII Cooking~\cite{rohrbach_ijcv_2015} has a larger number of videos but some actions are rarely performed.
Specifically, seven actions are performed fewer than ten times each. Furthermore there is gratuitous use of a background class because it was labeled for (sparse) action detection instead of (dense) action segmentation.
%which results in an insufficient ratio of instances per action. 
%For good performance we may need to replace our semi-Markov model, which works well for (dense) segmentation, with a detection model.
%Furthermore, the common evaluation setup is meant for action detection, not segmentation and classification. 
Georgia Tech Egocentric Activities~\cite{fathi_iccv_2011} has 28 videos across seven tasks. Unfortunately, the actions in each task are independent thus there are only three videos to train on and one for evaluation. 
Furthermore the complexities of egocentric video are beyond the scope of this work.
We use datasets from the ubiquitous computing and surgical robotics communities which contain many instances of each action.

\fakesubsection{University of Dundee 50 Salads}
Stein and McKenna introduced 50 Salads \cite{stein_ubicomp_2013} for evaluating fine-grained action recognition in the cooking domain. 
%Trials include video, depth maps (via Kinect), and synchronized 3-axis accelerometer data from 10 kitchen objects. 
We believe this dataset provides great value to the computer vision community due to the large number of action instances per class, the high quality labels, plethora of data, and multi-modal sensors (RGB, depth, and accelerometers). 
%It can be used to answer many research questions regarding fine-grained action analysis. 

%These accelerometers can be used as a proxy for which objects are being used in the scene. 
%The objects are: \textit{plate}, \textit{pepper dispenser}, \textit{bowl}, \textit{oil bottle}, \textit{large spoon}, \textit{dressing glass}, \textit{knife}, \textit{peeler}, \textit{small spoon}, and \textit{chopping board}.

This dataset includes 50 instances of salad preparation, where each of the 25 users makes a salad in two different trials. 
Videos are annotated at four levels of granularity. The coarsest level (``high'') consists of labels \textit{cut and mix ingredients}, \textit{prepare dressing}, and \textit{serve salad}. At the second tier (``mid'') there are 17 fine-grained actions like \textit{add vinegar}, \textit{cut tomato}, \textit{mix dressing}, \textit{peel cucumber}, \textit{place cheese into bowl}, and \textit{serve salad}. At the finest level (``low'') there are 51 actions indicating the start, middle, and end of the previous 17 actions. For each granularity there is also a \textit{background} class. 
%All actions are listed in the supplemental material. 

%The second granularity (``eval") adds object-agnostic actions like \textit{cutting} 
%%consolidates object-specific actions like \textit{cutting a tomato} and \textit{cutting a cucumber} into a . 
%%For example instead of having a separate action for \textit{cutting a cucumber} and \textit{cutting cheese} there is a single \textit{cutting} action. 
%These actions are: \textit{add dressing, add oil, add pepper, cut, mix dressing, mix ingredients, peel, place} and \textit{serve salad}.

A fourth granularity (``eval"), suggested by \cite{stein_ubicomp_2013}, consolidates some object-specific actions like \textit{cutting a tomato} and \textit{cutting a cucumber} into object-agnostic actions like \textit{cutting}.
%For example instead of having a separate action for \textit{cutting a cucumber} and \textit{cutting cheese} there is a single \textit{cutting} action. 
Actions include \textit{add dressing}, \textit{add oil}, \textit{add pepper}, \textit{cut}, \textit{mix dressing}, \textit{mix ingredients}, \textit{peel}, \textit{place}, \textit{serve salad on plate}, and \textit{background}. These labels coincide with the tools instrumented with accelerometers.

\fakesubsection{JHU-ISI Gesture and Skill Assessment Working Set (JIGSAWS)}\\
JIGSAWS \cite{JIGSAWS} was developed for recognizing actions in robotic surgery training tasks like suturing, needle passing, and knot tying. 
In this work we evaluate using the suturing task, which includes 39 trials of synchronized video and robot kinematics data collected from a daVinci medical robot. The video is captured from an overhead endoscopic camera and depicts two tools and the training task apparatus. 
%There are eight users and five trials per user. 
The suturing task consists of 10 fine-grained actions such as \textit{insert needle into skin}, \textit{tie a knot}, \textit{transfer needle between tools}, and \textit{drop needle at finish}.
Videos last about two minutes and contain 15 to 37 action instances per video. 
Users perform low-level actions in significantly different orders.
We evaluate using Leave One User Out as described in \cite{JIGSAWS}. 
Most prior work on this dataset focuses on the kinematics data which consists of positions, velocities, and robot joint information. We compare against the video-based results of Tao \etal~\cite{tao_miccai_2013}, which uses holistic features with a Markov Semi-Markov CRF.

%In the sensor-based baseline we use the robot kinematics and vision-based features from the authors of \cite{lea_wacv_2015}. The features are: left and right tool positions, velocities, and gripper angles as well as the distance from the tools to the closest object in the scene from the video.

%------------------------------------------------------------------------------------
\fakesubsection{Metrics}
We evaluate on segmental and frame-wise metrics as suggested by \cite{lea_icra_2016} for the 50 Salads dataset. The first measures segment \textit{cohesion} and the latter captures overall \textit{coverage}. 
%and the other measures overall prediction accuracy. 

The segmental metric evaluates the ordering of actions but not the specific timings. The motiviation is that in many applications there is high uncertainty in the location of temporal boundaries. For example, different annotators may have different interpretations of when an action starts or ends. 
%For example, should \textit{cutting} start when a user reaches for the knife or after the knife is in hand? 
As such, the precise location of temporal boundaries may be inconsequential. 
%In other applications, like human-robot interaction, having cohesive segments is more important. For example, if predictions oscillate between two states in a short amount of time (e.g. $AABABAAA$), despite having a high overall accuracy, the system may not function properly. These over segmentation errors should be penalized.
%The segmental edit score, $A_{edit}(P,P^*)$, that allows for slight temporal deviations between the ground truth and predictions but penalizes over segmentation.
This score, $A_{edit}(P,P^*)$, is computed using the Levenshtein distance, which is a function of segment insertions, deletions, and substitutions~\cite{Navarro_2001}. Let the ground truth segments be $P=\{P_1,\dots, P_M\}$ and predicted segments be $P^*=\{P^*_1,\dots, P^*_N \}$. The number of edits is normalized by the maximum of $M$ and $N$. For clarity we show the score $(1-A_{edit}(P,P^*)) \times 100 $, which ranges from 0 to 100. 

Frame-wise accuracy measures the percentage of correct frames in a sequence. 
%and is useful for determining how This is important for coverage. 
%coverage
%Due to issues in inter-annotator variability the score may be sufficient 
Let $y=\{y_1,\dots,y_T\}$ be the true labels and $y^*=\{y^*_1,\dots,y^*_T\}$ be the predicted labels. The score is a function of each frame: $A_{acc}(y,y^*)=\frac{1}{T} \sum_{t=1}^T \mathbf{1}(y_t=y^*_t)$. 

We also include action classification results which assume temporal segmentation is known and compare with the video-based results from Zappella \etal~\cite{zappella_media_2013}. These use the accuracy metric applied to segments instead of individual frames. 

\fakesubsection{Baselines} We evaluate two spatial baselines on both datasets using IDT and a pre-trained VGG network, and one temporal baseline using an RNN with LSTM. 
For the classification results, the (known) start and end times are fed into the segmental model to predict each class.

The IDT baseline is comparable to Rohrbach \etal~\cite{rohrbach_ijcv_2015} on the MPII dataset. We extract IDT, create a KMeans dictionary ($k=2000$), and aggregate the dictionary elements into a locally normalized histogram with a sliding window of 30 frames. 
We only use one feature type, HOG, because it outperformed all other feature types or their combination.
%Bojanowski \etal \cite{bojanowski_eccv_2014} found similar results.
This may be due to the large dimensionality of IDT and relatively low number of samples from our training sets. 
%For validation we evaluated the IDT baseline on Georgia Tech Egocentric Activities~\cite{?} and achieved comparable performance to Li \etal~\cite{li_cvpr_2015}. 
Note that it took 18 hours to compute IDT features on 50 Salads compared to less than 5 hours for the CNN features using an Nvidia Titan X graphics card.
%\TODO{mention our IDT results are comparable to Li on GTEA}

For our spatial-only results, we classify the action at each time step with a linear Support Vector Machine using the features from IDT, VGG, or our spatial CNN. These results highlight how effective each model is at representing the scene and are not meant to be state of the art.  
The CNN baseline uses the VGG network~\cite{VGG} pretrained on Imagenet. 
We use the activations from FC6, the first of VGG's three fully connected layers, as the features at each frame.

In addition we compare our temporal model to an RNN with LSTM using our spatial CNN as input. 
The LSTM baseline was implemented in Keras and uses one LSTM layer with 64 latent states.

\section{Results \& Discussion}
\label{sec:discussion}

%-------------------------------------------------------------
%-------------------------------------------------------------
\setcounter{table}{0}
\begin{table}[t]
	%\begin{center}
	\begin{minipage}[t]{0.5\linewidth}
		\begin{center}
			%                   \textbf{50 Salads} \\
			\begin{tabular}{| l | c | c |}
				\hline
				\textbf{Spatial Models}  & \textbf{Edit} & \textbf{Accuracy}\\
				\hline              
				VGG   & 7.58 & 38.30 \\ 				
				IDT   & 16.77 & 54.28 \\   
				S-CNN  & \textbf{24.10} & \textbf{66.64}\\
				%				IDT+LSTM   & ? & ? \\
				%				IDT+Evo   & ? & 61.46 \\
				%				\hline
				%				\textbf{Our model}  & \textbf{Edit} & \textbf{Accuracy}\\
				%				\hline      
				\hline 
				\textbf{Spatiotemporal}  & \textbf{Edit} & \textbf{Accuracy}\\	
				\hline								        
				S-CNN + LSTM  & 58.84 & 66.30 \\ 
				% clf = 74.63
				%                       Scene+LSTM+Seg  & 61.43 & 66.33 \\                      
				ST-CNN  & 60.98 & 71.37 \\ 
				ST-CNN + Seg  &\textbf{62.06} & \textbf{72.00}\\ 
				\hline
			\end{tabular}
			\caption{50 Salads (``eval'' granularity)}
			\label{table:salads}
		\end{center}
	\end{minipage}
	%            &
	\begin{minipage}[t]{0.5\linewidth}
		\begin{center}
			%                   \textbf{JIGSAWS} \\
			\begin{tabular}{| l | c | c |}
				\hline
				\textbf{Spatial Models}  & \textbf{Edit} & \textbf{Accuracy}\\
				\hline          
				VGG 		& \textbf{24.29} & 45.91 \\
				IDT           & 8.45 & 53.92 \\
				%				IDT+LSTM  & ? & ?\\
				%               sCNN (framewise)  & ? & 56.29 \\ 
				%				IDT+Evo    & ? & 72.10
				S-CNN  & 20.10 & \textbf{67.25}\\
				\hline 
				\textbf{Spatiotemporal}  & \textbf{Edit} & \textbf{Accuracy}\\				
				\hline				
				\cite{tao_miccai_2013} STIPS+CRF  & -& 71.78 \\                 
				%				\hline
				%				\textbf{Our model}  & \textbf{Edit} & \textbf{Accuracy}\\
%				\hline
				S-CNN + LSTM  & 54.07 & 68.37\\               
				%                       Scene+LSTM+Seg  & 58.71 & 68.77\\
				ST-CNN &  59.89 & 71.21\\
				ST-CNN + Seg &  \textbf{66.56} & \textbf{74.22}\\
				% clf=90.47
				\hline
			\end{tabular}
		\end{center}
		\caption{JIGSAWS}
		\label{table:jigsaws}
	\end{minipage}
	\\
	\\
	\\
	\begin{minipage}[t]{0.5\linewidth}
		\begin{center}      
			%           \textbf{50 Salads Granularities}\\              
			%   \begin{tabular}{cc}
			% -----------Hierarchy----------
			\begin{tabular}{| l | c |c | c || c |}
				%       \hline
				%       \textbf{Sensors} & \textbf{Classes} & \textbf{Edit} & \textbf{Acc.}  & \textbf{Classif.} \\
				%       \hline
				%       Low & 52  & 30.19 & 50.37 &  26.86\\
				%       Mid   & 18 &  36.10 & 58.33 &   44.03\\
				%       Eval   & 10 &  55.50 & 82.14 &   64.32\\
				%       High  &  4 &  61.60 & 94.48  &  69.03\\
				\hline
				\textbf{Labels} & \textbf{Classes} & \textbf{Edit} & \textbf{Acc.}  & \textbf{Classif.} \\
				\hline
				% =====ICRA results=====    
				%       icra (saturday)  & & & & \\
				Low & 52   & 29.30  & 44.13 & 39.67 \\
				Mid   & 18 &  48.94 & 58.06  & 63.49 \\
				Eval  & 10 & 62.06  & 72.00 &  86.63\\
				High & 4 & 83.2  & 92.43 & 95.14 \\     
				%       \hline
				%       icra (saturday)  & & & & \\
				%       Low & 52   & 29.30  & 44.13 & 39.67 \\
				%       Mid   & 18 &  48.94 & 58.06  & 63.49 \\
				%       Eval  & 10 & 66.44  & 72.71 &  86.63\\
				%       High & 4 & 83.2  & 92.43 & 95.14 \\
				\hline
				%       old & & & & \\
				%       % =====Results from CVPR=====       
				%       Low & 52   &  32.83 &  37.90&  35.56\\
				%       Mid   & 18 &  43.17 & 56.93 &  55.07\\
				%       Eval  & 10 &  55.96 & 69.36 &   69.06\\
				%       High & 4 &  62.85 & 91.54 & 87.50\\
				%       \hline
			\end{tabular}
			\caption{50 Salads Granularity Analysis}
			\label{table:granularities}
		\end{center}        
	\end{minipage}
	\begin{minipage}[t]{0.5\linewidth}
		\begin{center}
			%       \textbf{Speedup Analysis}\\
			\begin{tabular}{| l |  c | c | c |}
				\hline
				\textbf{Labels} & \textbf{Dur} & \textbf{\#Segs} & \textbf{Speedup} \\      
				\hline
				Low &  2289 & 65 & 35x\\ 
				Mid &  3100 & 25 & 124x \\ 
				Eval & 3100 & 24 & 129x \\ 
				High & 11423 & 6 & 1902x\\ 				
%				Low &  2289 & 83 & 28x\\ 
%				Mid &  3100 & 40 & 85x \\ 
%				Eval & 3100 & 37 & 84x \\ 
%				High & 11423 & 10 & 1142x\\ 
				\hline
				JIGSAWS & 1107 & 37 & 30x\\                 
				\hline          
			\end{tabular}
			\caption{Speedup Analysis}
			\label{table:speedup}
		\end{center}            
	\end{minipage}      
\end{table}

\begin{figure*}[t]
	\includegraphics[width=\hsize]{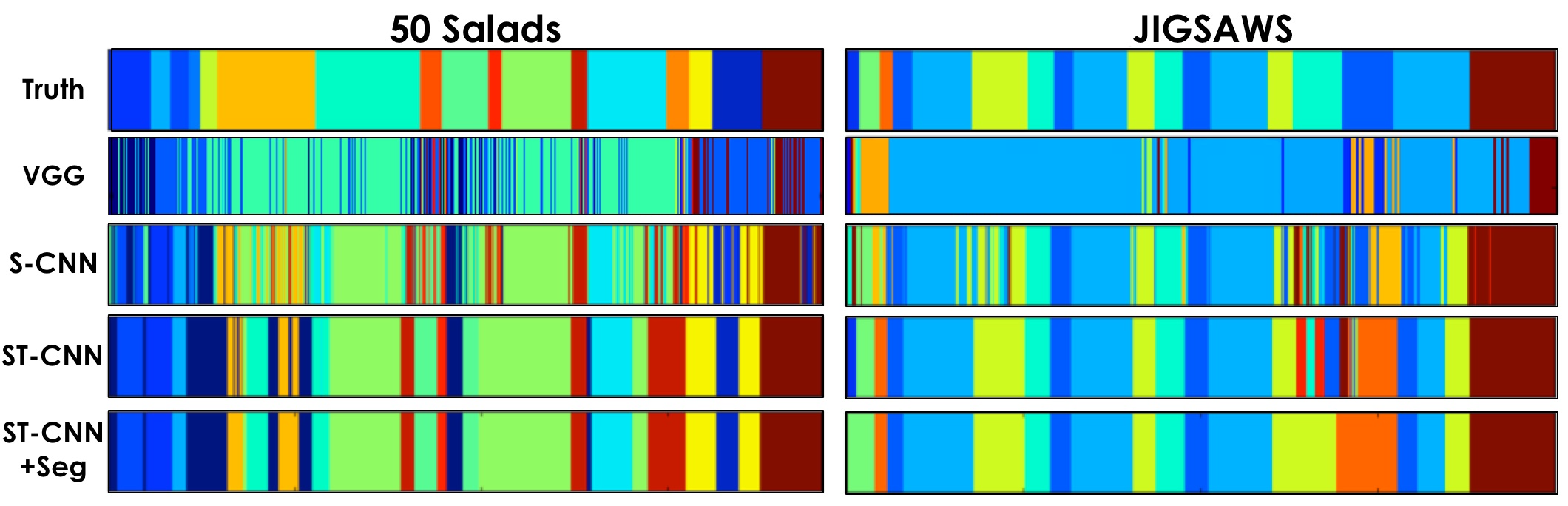}
	\caption{The plots on top depict the ground truth action predictions for a given video. Each color corresponds to a different class label. Subsequent rows show predictions using VGG, S-CNN, ST-CNN, and ST-CNN + Seg. 
%		The second row, S-CNN, depicts the predictions using the spatial component of our model. The third row, ST-CNN, depicts predictions using the spatiotemporal model with frame-wise inference. The bottom row, ST-CNN + Seg uses the spatiotemporal model with segmental inference.
		}
	\label{fig:timeline}
\end{figure*}

%\fakesubsection{Results}
%-------------------------------------------------------------
%-------------------------------------------------------------
%The results in Tables~\ref{table:salads} and \ref{table:jigsaws} 
%Regardless of the temporal model, performance of our CNN-based scene model is superior to the IDT baselines.
Tables~\ref{table:salads} and \ref{table:jigsaws} show performance using IDT, VGG, LSTM, and our models and Figure~\ref{fig:timeline} shows example predictions on each dataset.
S-CNN, ST-CNN, and ST-CNN + Seg refer to the spatial, spatiotemporal, and segmental components of our model. 
%These 50 Salads results are on the ``eval" granularity. 
Our full model has 27.8\% better accuracy on 50 Salads and 37.6\% better accuracy on JIGSAWS relative to the IDT baseline.

\fakesubsection{Spatial Model}
Our results are consistent with the claim that holistic methods like IDT are insufficient for fine-grained action segmentation. 
Interestingly, we also see that the VGG results are also relatively poor, which could be due to the data augmentation to train the model. 
%That is not to say that our results are sufficient, but 
%For the fairest comparison attend to the scores for our scene component versus IDT.
While our results are still insufficient for many practical applications the accuracy of our spatial model is at least 12\% better than IDT and 21\% better than VGG on both datasets. 
Note that the edit score is very low for all of these models.
This is not surprising because each model only uses local temporal information, which results in many oscillations in predictions, as shown in Figure~\ref{fig:timeline}.
% predictions from the spatial model tends to oscillate over time. 

Many actions in 50 Salads, like cutting, require capturing small hand motions. We visualized IDT\footnote{Visualization was performed using the public software from Wang \etal~\cite{wang_iccv_2013}.}
and found it does not detect many tracklets for these actions. In contrast, when the user places ingredients in the bowl IDT generates thousands of tracklets. 
%We found poor performance even when compensating for the imbalance by normalizing our feature histograms. 
We found this to be problematic despite the fact that the IDT features are normalized. 
Qualitatively found our model is better at capturing details necessary for finer motions, as shown in Figure~\ref{fig:activations}. 

\fakesubsection{Temporal Model}
%Based on ablative analysis of our model we see that each component makes an important contribution. 
The spatiotemporal model (ST-CNN) outperforms the spatial model (S-CNN) on both datasets. 
The effect on edit score is substantial and likely due to the large temporal filters. Aside from modeling temporal evolution these have a byproduct of smoothing out predictions. By visualizing these features we see they tend to capture different phases of an action like the start or finish.
In contrast, while LSTM substantially improves edit score over the spatial model it has a negligible impact on accuracy. 
LSTM is capable of learning how actions transition across time, however, it does not appear that it sufficiently captures this information. Due to the complex nature of this method, we were not able to visualize the internal parameters in a meaningful way.

\fakesubsection{Segmental Model}
The segmental model provides a notable improvement on JIGSAWS but only a modest improvement on 50 Salads. 
By visualizing the results we see that the segmental model helps in some cases and hurts in others. For example, when the predictions oscillate (like in Figure~\ref{fig:timeline} (right)) the segmental model provides a large improvement. However, sometimes the model smooths over actions that are short in duration. Future work should look at incorporating additional cues such as duration to better model each action class.

%In many cases edit performance improves by 10\%. 
%For example on JIGSAWS the \textit{Scene+Evo} model achieves 71.23\% and \textit{Scene+Evo+Seg} achieves 82.16\%. 
%Interestingly, in most cases the accuracy between frame-wise and segmental does not vary significantly. 
%This implies that the percent of frames that are correct is roughly constant but the number of false-positives decreases. This is important for robotics applications where false-positives can be especially problematic. \TODO{elaborate}

\fakesubsection{Action Granularity} Table~\ref{table:granularities} shows performance on all four action granularities from 50 Salads using our full model. Columns 3 and 4 show scores for segmental and frame-wise metrics on the action segmentation task and the last shows action classification accuracies assuming known temporal segmentation. 
%While the performance decreases going from ``high" to ``low" the difference in sensor accuracies for ``medium" (18 classes) compared to ``low" (52 action classes) is relatively small. 
While performance decreases as the number of classes increases, results degrade sub-linearly with each additional class. Some errors at the finer levels are likely due to temporal shifts in the predictions. 
Given the high accuracy at the coarser levels, future work should look at hierarchical modeling of finer granularities.
%The classification scores are modest. 
%This is in large part due to the background class. There are many very short background actions that happen in between other long segments. Our model does not recognize these well which has a large impact on classification accuracy. 

\fakesubsection{Other Results}
Lea \etal~\cite{lea_icra_2016} achieved an edit score of 58.46\% and accuracy of 81.75\% using the instrumented kitchen tools on 50 Salads. 
They also achieved state of the art performance on JIGSAWS with 78.91\% edit and 83.45\% accuracy.
These results used domain-specific sensors which are well suited to each application but may not be practical for real-world deployment. To contrast, video is much more practical for deployment but is more complicated to model. 
Therefore, we should not expect to achieve as high performance from video.

Our classification accuracy on JIGSAWS is 90.47\%. This is notably higher than the state of the art~\cite{zappella_media_2013}, which achieved 81.17\% using a video-based linear dynamical system model and also better than their hybrid approach using video and kinematics, which achieved 86.56\%.
For joint segmentation and classification the improvement over the state of the art~\cite{tao_miccai_2013} is modest. 
These surgical actions can be recognized well using position and velocity information~\cite{lea_icra_2016}, thus our ability to capture object relationships may be less important on this dataset. 

\fakesubsection{Speedup}
Table~\ref{table:speedup} shows the speedup of our inference algorithm compared to Segmental Viterbi on all 50 Salads and JIGSAWS label sets.
One practical implication is that our algorithm scales readily to full-length videos. 

\section{Conclusion}
\label{sec:conclusion}
In this paper we introduced a segmental spatiotemporal CNN that substantially outperforms popular methods like Dense Trajectories, pre-trained spatial CNNs, and temporal models like RNNs with LSTM. 
Furthermore, our approach takes less time to compute features than IDT, less time to train than LSTM, and performs inference more efficiently than traditional Segmental methods.
%We hope the insights from our discussions are useful for highlighting the nuances of fine-grained action recognition. 

\fakesubsection{Acknowledgments} This work was funded in part by grants NIH R01HD87133-01,
ONR N000141310116, and an Intuitive Surgical Research Grant.

\clearpage

\bibliographystyle{splncs}
%\bibliography{egbib}

\bibliography{bib/activity_recognition,bib/surgical}

\begin{thebibliography}{10}

\bibitem{wang_iccv_2013}
Wang, H., Schmid, C.:
\newblock Action recognition with improved trajectories.
\newblock In: IEEE International Conference on Computer Vision (ICCV). (2013)

\bibitem{sun_iccv_2015}
Sun, L., Jia, K., Yeung, D.Y., Shi, B.:
\newblock Human action recognition using factorized spatio-temporal
  convolutional networks.
\newblock In: IEEE International Conference on Computer Vision (ICCV). (2015)

\bibitem{heilbron_cvpr_2015}
Caba~Heilbron, F., Escorcia, V., Ghanem, B., Carlos~Niebles, J.:
\newblock {ActivityNet}: A large-scale video benchmark for human activity
  understanding.
\newblock In: IEEE Conference on Computer Vision and Pattern Recognition
  (CVPR). (2015)

\bibitem{rohrbach_ijcv_2015}
Rohrbach, M., Rohrbach, A., Regneri, M., Amin, S., Andriluka, M., Pinkal, M.,
  Schiele, B.:
\newblock Recognizing fine-grained and composite activities using hand-centric
  features and script data.
\newblock International Journal of Computer Vision (IJCV) (2015)

\bibitem{fathi_cvpr_2013}
Fathi, A., Rehg, J.M.:
\newblock Modeling actions through state changes.
\newblock In: IEEE Conference on Computer Vision and Pattern Recognition
  (CVPR). (2013)

\bibitem{li_cvpr_2015}
Li, Y., Ye, Z., Rehg, J.M.:
\newblock Delving into egocentric actions.
\newblock In: IEEE Conference on Computer Vision and Pattern Recognition
  (CVPR). (2015)

\bibitem{cheron_iccv_2015}
Cheron, G., Laptev, I., Schmid, C.:
\newblock P-cnn: Pose-based cnn features for action recognition.
\newblock In: IEEE International Conference on Computer Vision (ICCV). (2015)

\bibitem{ni_cvpr_2014}
Ni, B., Paramathayalan, V.R., Moulin, P.:
\newblock Multiple granularity analysis for fine-grained action detection.
\newblock In: IEEE Conference on Computer Vision and Pattern Recognition
  (CVPR). (2014)

\bibitem{vo_cvpr_2014}
Vo, N.N., Bobick, A.F.:
\newblock From stochastic grammar to bayes network: Probabilistic parsing of
  complex activity.
\newblock In: IEEE Conference on Computer Vision and Pattern Recognition
  (CVPR). (2014)

\bibitem{hawkins_icra_2014}
Hawkins, K.P., Bansal, S., Vo, N.N., Bobick, A.F.:
\newblock Anticipating human actions for collaboration in the presence of task
  and sensor uncertainty.
\newblock In: IEEE International Conference on Robotics and Automation (ICRA).
  (2014)

\bibitem{JIGSAWS}
Gao, Y., Vedula, S.S., Reiley, C.E., Ahmidi, N., Varadarajan, B., Lin, H.C.,
  Tao, L., Zappella, L., B{\'e}jar, B., Yuh, D.D.,  et~al.:
\newblock {JHU-ISI Gesture and Skill Assessment Working Set} (jigsaws): A
  surgical activity dataset for human motion modeling.
\newblock MICCAI Workshop: Modeling and Monitoring of Computer Assisted
  Interventions (M2CAI) (2014)

\bibitem{zappella_media_2013}
Zappella, L., Haro, B.B., Hager, G.D., Vidal, R.:
\newblock Surgical gesture classification from video and kinematic data.
\newblock Medical Image Analysis (2013)

\bibitem{tao_miccai_2013}
Tao, L., Zappella, L., Hager, G.D., Vidal, R.:
\newblock Surgical gesture segmentation and recognition.
\newblock In: Medical Image Computing and Computer Assisted Intervention
  (MICCAI). (2013)  339--346

\bibitem{lei_ubicomp_2012}
Lei, J., Ren, X., Fox, D.:
\newblock Fine-grained kitchen activity recognition using {RGB-D}.
\newblock In: ACM Conference on Ubiquitous Computing (UbiComp). (2012)

\bibitem{morariu_cvpr_2011}
Morariu, V.I., Davis, L.S.:
\newblock Multi-agent event recognition in structured scenarios.
\newblock In: IEEE Conference on Computer Vision and Pattern Recognition
  (CVPR). (2011)

\bibitem{koppula_ijrr_2013}
Koppula, H.S., Gupta, R., Saxena, A.:
\newblock Learning human activities and object affordances from rgb-d videos.
\newblock The International Journal of Robotics Research (IJRR) (2013)

\bibitem{van_jaise_2010}
van Kasteren, T., Englebienne, G., Kr{\"o}se, B.J.:
\newblock Activity recognition using semi-markov models on real world smart
  home datasets.
\newblock Journal of Ambient Intelligence and Smart Environments (2010)

\bibitem{VGG}
Simonyan, K., Zisserman, A.:
\newblock Very deep convolutional networks for large-scale image recognition.
\newblock In: International Conference Learning Representations (ICLR). (2015)

\bibitem{alexnet}
Krizhevsky, A., Sutskever, I., Hinton, G.E.:
\newblock Imagenet classification with deep convolutional neural networks.
\newblock In Pereira, F., Burges, C., Bottou, L., Weinberger, K., eds.:
  Advances in Neural Information Processing Systems (NIPS).
\newblock (2012)

\bibitem{sarawagi_nips_2004}
Sarawagi, S., Cohen, W.W.:
\newblock Semi-markov conditional random fields for information extraction.
\newblock In: Advances in Neural Information Processing Systems (NIPS). (2004)

\bibitem{stein_ubicomp_2013}
Stein, S., McKenna, S.J.:
\newblock Combining embedded accelerometers with computer vision for
  recognizing food preparation activities.
\newblock In: ACM Conference on Ubiquitous Computing (UbiComp). (2013)

\bibitem{wang_cvpr_2011}
Wang, H., Kl{\"a}ser, A., Schmid, C., Liu, C.L.:
\newblock {Action Recognition by Dense Trajectories}.
\newblock In: IEEE Conference on Computer Vision and Pattern Recognition
  (CVPR). (2011)

\bibitem{pirsiavash_cvpr_2014}
Pirsiavash, H., Ramanan, D.:
\newblock Parsing videos of actions with segmental grammars.
\newblock In: IEEE Conference on Computer Vision and Pattern Recognition
  (CVPR). (2014)

\bibitem{jain_cvpr_2015}
Jain, M., van Gemert, J.C., Snoek, C.G.M.:
\newblock What do 15,000 object categories tell us about classifying and
  localizing actions?
\newblock In: IEEE Conference on Computer Vision and Pattern Recognition
  (CVPR). (2015)

\bibitem{pishchulin_gcpr_2014}
Pishchulin, L., Andriluka, M., Schiele, B.:
\newblock Fine-grained activity recognition with holistic and pose based
  features.
\newblock In: German Conference on Pattern Recognition (GCPR). (2014)

\bibitem{karpathy_cvpr_2014}
Karpathy, A., Toderici, G., Shetty, S., Leung, T., Sukthankar, R., Fei-Fei, L.:
\newblock Large-scale video classification with convolutional neural networks.
\newblock IEEE Conference on Computer Vision and Pattern Recognition (CVPR)
  (2014)

\bibitem{simonyan_nips_2014}
Simonyan, K., Zisserman, A.:
\newblock Two-stream convolutional networks for action recognition in videos.
\newblock In: Advances in Neural Information Processing Systems (NIPS). (2014)
  568--576

\bibitem{tran_iccv_2015}
Tran, D., Bourdev, L., Fergus, R., Torresani, L., Paluri, M.:
\newblock Learning spatiotemporal features with 3d convolutional networks.
\newblock In: The IEEE International Conference on Computer Vision (ICCV).
  (2015)

\bibitem{peng_thumos_2015}
Peng, X., Schmid, C.:
\newblock Encoding feature maps of cnns for action recognition.
\newblock In: CVPR, THUMOS Challenge 2015 Workshop. (2015)

\bibitem{ng_cvpr_2015}
Ng, J.Y., Hausknecht, M.J., Vijayanarasimhan, S., Vinyals, O., Monga, R.,
  Toderici, G.:
\newblock Beyond short snippets: Deep networks for video classification.
\newblock In: IEEE Conference on Computer Vision and Pattern Recognition
  (CVPR). (2015)

\bibitem{shi_ijcv_2011}
Shi, Q., Cheng, L., Wang, L., Smola, A.:
\newblock Human action segmentation and recognition using discriminative
  semi-markov models.
\newblock International Journal of Computer Vision (IJCV) (2011)

\bibitem{tang_cvpr_2012}
Tang, K., Fei-Fei, L., Koller, D.:
\newblock Learning latent temporal structure for complex event detection.
\newblock In: IEEE Conference on Computer Vision and Pattern Recognition
  (CVPR). (2012)

\bibitem{Vinyals_2015_CVPR}
Vinyals, O., Toshev, A., Bengio, S., Erhan, D.:
\newblock Show and tell: A neural image caption generator.
\newblock In: IEEE Conference on Computer Vision and Pattern Recognition
  (CVPR). (2015)

\bibitem{lrcn2014}
Donahue, J., Hendricks, L.A., Guadarrama, S., Rohrbach, M., Venugopalan, S.,
  Saenko, K., Darrell, T.:
\newblock Long-term recurrent convolutional networks for visual recognition and
  description.
\newblock In: IEEE Conference on Computer Vision and Pattern Recognition
  (CVPR). (2015)

\bibitem{ng_2015}
Hannun, A.Y., Case, C., Casper, J., Catanzaro, B.C., Diamos, G., Elsen, E.,
  Prenger, R., Satheesh, S., Sengupta, S., Coates, A., Ng, A.Y.:
\newblock Deep speech: Scaling up end-to-end speech recognition.
\newblock CoRR \textbf{abs/1412.5567} (2014)

\bibitem{abdel_interspeech_2013}
Abdel{-}Hamid, O., Deng, L., Yu, D.:
\newblock Exploring convolutional neural network structures and optimization
  techniques for speech recognition.
\newblock In: {INTERSPEECH} International Speech Communication Association.
  (2013)

\bibitem{lea_icra_2016}
Lea, C., Vidal, R., Hager, G.D.:
\newblock Learning convolutional action primitives for fine-grained action
  recognition.
\newblock In: IEEE International Conference on Robotics and Automation (ICRA).
  (2016)

\bibitem{Goodfellow_book}
Ian~Goodfellow, Y.B., Courville, A.:
\newblock Deep learning.
\newblock Book in preparation for MIT Press (2016)

\bibitem{davis_cvpr_1997}
Davis, J., Bobick, A.:
\newblock The representation and recognition of action using temporal
  templates.
\newblock In: IEEE Conference on Computer Vision and Pattern Recognition
  (CVPR). (1997)

\bibitem{ADAM}
Kingma, D.P., Ba, J.:
\newblock Adam: {A} method for stochastic optimization.
\newblock In: International Conference Learning Representations (ICLR). (2014)

\bibitem{keras}
Chollet, F.:
\newblock Keras.
\newblock \url{https://github.com/fchollet/keras} (2015)

\bibitem{fathi_iccv_2011}
Fathi, A., Farhadi, A., Rehg, J.M.:
\newblock Understanding egocentric activities.
\newblock In: IEEE International Conference on Computer Vision (ICCV). (2011)

\bibitem{Navarro_2001}
Navarro, G.:
\newblock A guided tour to approximate string matching.
\newblock ACM Comput. Surv. (2001)

\end{thebibliography}

\end{document}